\pdfoutput=1

\documentclass[11pt]{article}

\usepackage[final]{acl}

\usepackage{times}
\usepackage{latexsym}
\usepackage{booktabs}
\usepackage{xcolor}

\usepackage[T1]{fontenc}

\usepackage[utf8]{inputenc}

\usepackage{amssymb}

\usepackage{microtype}

\usepackage{inconsolata}

\usepackage{graphicx}

\usepackage{pgfplots}

\usepackage{multirow}
\usepackage{subcaption}

\usepackage{maltese}

\usepackage{arabtex}
\usepackage{utf8}

\usepackage{stackengine}
\usepackage{tikz}
\newcommand\dottedcircle{\tikz \draw [line cap=round, line width=0.15ex, dash pattern=on 0pt off 1.95\pgflinewidth] (0,0) circle [radius=0.6ex];}
\makeatletter
\newcommand{\fatha}{%
\bgroup \set@arabfont
\stackon[0.5pt]{\dottedcircle}{\char'013}%
\egroup%
}
\newcommand{\kasra}{%
\bgroup \set@arabfont
\stackunder[0.5pt]{\dottedcircle}{\char'013}%
\egroup%
}
\newcommand{\damma}{%
\bgroup \set@arabfont
\stackon[0.5pt]{\dottedcircle}{\char'014}%
\egroup%
}
\newcommand{\fathatan}{%
\bgroup \set@arabfont
\stackon[0.5pt]{\dottedcircle}{\char'023}%
\egroup%
}
\newcommand{\kasratan}{%
\bgroup \set@arabfont
\stackunder[0.5pt]{\dottedcircle}{\char'023}%
\egroup%
}
\newcommand{\shaddah}{%
\bgroup \set@arabfont
\stackon[0.5pt]{\dottedcircle}{\char'017}%
\egroup%
}
\newcommand{\dammatan}{%
\bgroup \set@arabfont
\stackon[0.5pt]{\dottedcircle}{\char'024}%
\egroup%
}
\newcommand{\skun}{%
\bgroup \set@arabfont
\stackon[0.5pt]{\dottedcircle}{\char'025}%
\egroup%
}
\newcommand{\alif}{%
\bgroup \set@arabfont
\stackon[0.5pt]{\dottedcircle}{\char'027}%
\egroup%
}
\newcommand{\alifwaslafatha}{%
\bgroup \set@arabfont
\stackon[0.5pt]{\<ٱ>}{\char'013}%
\egroup%
}
\newcommand{\alifwaslakasra}{%
\bgroup \set@arabfont
\stackunder[0.5pt]{\<ٱ>}{\smash{\raisebox{1ex}{\char'013}}}%
\egroup%
}
\newcommand{\alifwasladamma}{%
\bgroup \set@arabfont
\stackon[0.5pt]{\<ٱ>}{\char'014}%
\egroup%
}
\makeatother

\setcode{utf8}

\newcommand{\hide}[1]{}

\newcommand{\AHAMZAUP}{{>}}

\title{Data Augmentation for Maltese NLP using\\Transliterated and Machine Translated Arabic Data}

\newcommand*{\authormark}[1][*]{\textsuperscript{#1}}
\author{
    Kurt Micallef\authormark[1,2]\\
    \texttt{kurt.micallef@um.edu.mt}\\\And
    Nizar Habash\authormark[2]\\
    \texttt{nizar.habash@nyu.edu}\\\And
    Claudia Borg\authormark[1]\\
    \texttt{claudia.borg@um.edu.mt}\\
    \AND %
    {\normalfont\authormark[1]Department of Artificial Intelligence, University of Malta}\\
    {\normalfont\authormark[2]Computational Approaches to Modeling Language Lab, New York University Abu Dhabi}\\
}

\begin{document}
\maketitle
\begin{abstract}
Maltese is a unique Semitic language that has evolved under extensive influence from Romance and Germanic languages, particularly Italian and English.
Despite its Semitic roots, its orthography is based on the Latin script, creating a gap between it and its closest linguistic relatives in Arabic.
In this paper, we explore whether Arabic-language resources can support Maltese natural language processing (NLP) through cross-lingual augmentation techniques.
We investigate multiple strategies for aligning Arabic textual data with Maltese, including various transliteration schemes and machine translation (MT) approaches.
As part of this, we also introduce novel transliteration systems that better represent Maltese orthography.
We evaluate the impact of these augmentations on monolingual and mutlilingual models and demonstrate that Arabic-based augmentation can significantly benefit Maltese NLP tasks.
\end{list} %
\end{abstract}

\section{Introduction}
\label{section:introduction}

Maltese is the only Semitic language written in the Latin script and the only one that is an official language of the European Union. It retains a core Semitic structure but has undergone extensive lexical borrowing and structural influence from Italian and English. Despite its roots in North African Arabic, modern Maltese and Arabic are now orthographically and lexically distant, posing unique challenges for leveraging Arabic NLP resources to support Maltese.

Given the low-resource status of Maltese, recent work has focused on leveraging multilingual language models and transfer learning (\citealp{lauscher-etal-2020-zero}; inter alia).
In this paper, we pursue a complementary strategy: enriching Maltese datasets through augmentation derived from Arabic resources.
While doing so, we address the divergence in script and phonology between Arabic and Maltese, by considering multiple layers of transliteration and translation to Arabic inputs.
These include transliteration schemes from previous works such as Buckwalter \cite{Buckwalter:2002:buckwalter}, Uroman \cite{hermjakob-etal-2018-box}, as well as machine translations.
As part of this work, we develop novel phonology- and morphology-aware transliteration systems, which we publicly release.\footnote{
\url{https://www.github.com/MLRS/maltify_arabic}
}
See example in Table~\ref{table:example}.
We apply these augmentations individually and in combination.

To assess the effectiveness of Arabic augmentation, we evaluate its impact across three language models: \textbf{mBERT}, \textbf{BERTu} (a Maltese BERT model), and \textbf{mBERTu} (mBERT with additional Maltese pre-training).
Our experiments demonstrate that the type of augmentation data significantly affects downstream performance and that models with more Maltese knowledge benefit differently from Arabic augmentation compared to less Maltese-aware models.
We also show that combining multiple augmentation techniques is helpful.

\begin{table}[t]
    \centering
    \renewcommand{\tabcolsep}{5pt}
\begin{tabular}{ll} 
    \toprule
    \textbf{Arabic} & \multicolumn{1}{r}{\<أوقفت السيارة في الطريق.>} \\
    \midrule
    \textbf{Buckwalter} & {\AHAMZAUP}wqft AlsyArp fy AlTryq. \\
    \textbf{Uroman} & awqft alsyara fy altryq. \\
    \textbf{CharTx} & uqft alsjara fi altriq. \\
    \textbf{MorphTx} & awqefat is-sejjara fit-teriq. \\
    \textbf{MT} & Waqqaft il-karozza fit-triq. \\
    \midrule
    \textbf{Maltese} & Ipparkjajt il-karozza fit-triq. \\
    \textbf{English} & I parked the car on the street. \\
    \bottomrule
\end{tabular}
    \caption{Arabic sentence and its Buckwalter, Uroman, and transliterations using our new systems (CharTx and MorphTx), along with Maltese machine translation (MT) and native Maltese versions.}
    \label{table:example}
\end{table}

The paper presents related work in Section~\ref{section:related}, describes our transliteration methods in Section~\ref{section:transliteration_systems}, outlines the experimental setup in Section~\ref{section:setup}, and presents our results in Sections~\ref{section:transliteration_results} and~\ref{section:cascaded_results}.

\section{Related Work}
\label{section:related}

\subsection{Cross-lingual Transfer}

Multilingual models have been shown to be quite effective for cross-lingual transfer \cite{wu-dredze-2019-beto}, but their effectiveness on low-resource languages is often limited by their representation in the model's pre-training data, which is often small or non-existent \cite{wu-dredze-2020-languages, lauscher-etal-2020-zero, muller-etal-2021-unseen, winata-etal-2022-cross}.
Low-resource languages are particularly limited to generalise well due to miniscule annotated datasets.
However, few-shot cross-lingual fine-tuning has been shown to be an effective strategy, where data from a high-resource language is used to improve performance on a low-resource language \cite{lauscher-etal-2020-zero, zhao-etal-2021-closer, schmidt-etal-2022-dont}.

Previous works have shown that script mismatch can particularly impact performance, while transliteration can be used as a way to align languages in a unified script and boost cross-lingual transfer \cite{muller-etal-2021-unseen, liu-etal-2025-transmi}.
Closer to our setting, \citet{micallef-etal-2023-exploring} study Maltese as a dialect of Arabic through transliteration.
We extend this by comparing broader data augmentation strategies and analysing their effects across models with varying degrees of Maltese similarity.

\subsection{Transliteration}

Various Arabic-to-Latin transliteration systems have been proposed with differing goals. Buckwalter maps Arabic letters to ASCII characters deterministically \cite{Buckwalter:2002:buckwalter}, while CAPHI emphasizes phonetic accuracy \cite{habash-etal-2018-unified}. Uroman offers general Latin-script mappings for many languages, including Arabic \cite{hermjakob-etal-2018-box}. \citet{eryani-habash-2021-automatic} propose a Romanisation system tailored for diacritised bibliographic records.
We use Buckwalter and Uroman as baselines in our experiments, but note that these representations often diverge from Maltese orthography, limiting their effectiveness for our task.

\section{Our Transliteration Systems}
\label{section:transliteration_systems}

In this section, we outline the two Arabic-to-Maltese transliteration systems we developed, with all mapping rules listed in Appendix~\ref{appendix:rules}.

\subsection{CharTx: Character Mappings}
\label{section:transliteration_character}

We define Arabic-to-Maltese character mappings by reversing the Maltese-to-Arabic rules from \citet{micallef-etal-2023-exploring}.
Ambiguities in that work, such as `s' mapping to \<س>~\textit{s} or~\<ص>~\textit{S}, collapse trivially to one Maltese form in our case.
More challenging are the Arabic glides \<ي>~\textit{y} and \<و>~\textit{w}, which can function as long vowels when preceded by  {\kasra}~\textit{i} and~{\damma}~\textit{u} diacritics, respectively \cite{habash-2010-arabic-nlp}.
We handle these by enumerating all diacritic combinations and mapping accordingly.

Our system avoids generating Maltese letters like `ċ', `g', `p', `v', and `z', as these largely arise from non-Arabic sources. For instance, while `g' could map from
\<ج>~\textit{j}, we consistently map it to `ġ', its more frequent counterpart. Letters like `p' and~`v' may relate to \<ب>~\textit{b} or \<ف>~\textit{f}, but are rarely found in words of Arabic origin.

Finally, characters not explicitly mapped are preserved.
The mappings are applied at the word level, so we tokenise using CAMeL Tools when needed \cite{obeid-etal-2020-camel}.
For cases where we tokenise we rejoin transliterated tokens and detokenise to remove unwanted spacing before punctuation.\footnote{
The original boundaries may not always be preserved.
}

\subsection{MorphTx: Morphological Features}
\label{section:transliteration_morphology}

On top of character mappings, we incorporate a linguistic disambiguator to predict diacritised forms using CAMeL Tools \cite{obeid-etal-2020-camel}, which implements the BERT-based model of \citet{inoue-etal-2022-morphosyntactic} with the Egyptian CALIMA C044 morphological database \cite{habash-etal-2012-morphological}.
We use the Egyptian model due to the greater similarity of Maltese to Dialectal Arabic over MSA.
Suitable analysers for closer dialects like Tunisian were not available.
However, when predictions fail under the Egyptian model, the system falls back to MSA.

The disambiguator selects the highest-scoring diacritised form from all possible morphological analyses in context, enabling more accurate application of the character mappings (CharTx) from Section~\ref{section:transliteration_character}. It also provides morpheme segmentation and POS tags, allowing us to define morpheme-specific mappings that override character-level rules.
For example, \texttt{DET}~\<ال>~\textit{Al} maps to `il-', and \texttt{PRON\_2MP} maps to `kom'. We include mappings for all Arabic affixes in the analyser database, and map them appropriately to Maltese.
We also capitalise morphemes tagged as proper nouns (\texttt{NOUN\_PROP}).
Additionally, following \citet{micallef-etal-2023-exploring}, we handle Maltese orthographic conventions such as the contraction of 
\textit{fi~il-} to \textit{fil-} `in the', and sun letter assimilation, e.g., \textit{il-żejt} to \textit{iż-żejt} `the oil'.

\section{Experimental Setup}
\label{section:setup}

Our goal is to improve cross-lingual transfer to Maltese by transliterating Arabic and using it for data augmentation.
Our few-shot setup involves fine-tuning on Arabic data, followed by Maltese.
Section~\ref{section:data} outlines the datasets used, and Section~\ref{section:models_inputs} details the models and input processing.
Appendix~\ref{appendix:fine_tuning} includes more technical details on our fine-tuning setup.

\subsection{Datasets}
\label{section:data}

\paragraph{Named-Entity Recognitions (NER)}
For Arabic, we use ANERCorp \cite{benajiba-etal-2007-anersys} with the splits from \citet{obeid-etal-2020-camel}.
For Maltese, we use the MAPA \cite{gianola-etal-2020-mapa} with the splits and fixes from \citet{micallef-etal-2024-cross}.
We normalise both datasets to a common tagset and also downsample the Maltese data.
After doing so, we have 3,973 and 155 training sentences for Arabic and Maltese, respectively.
Span-level F1 is used when evaluating model outputs.

\paragraph{Sentiment Analysis (SA)}
For Arabic we make use of the data from \citet{baly-etal-2018-arsentdlev}, while for Maltese we use the data from \citet{martinez-garcia-etal-2021-evaluating}.
Since the Maltese data only has positive and negative sentences, we drop any neutral sentences from the Arabic data.
After this filtering process we have 15,305 and 595 training sentences for Arabic and Maltese, respectively.
We use macro-averaged F1 to evaluate the models.

All of the Arabic data is pre-processed using CAMeL Tools \texttt{arclean}, which normalises ambiguous Arabic characters, that could be potentially problematic in our modelling \cite{obeid-etal-2020-camel}.
Appendix~\ref{appendix:data_normalisation} includes more details on the datasets used, including our filtering and processing steps.
The Arabic data is only used for training, and the validation and testing data are always Maltese.
Table~\ref{table:data_sizes} provides a breakdown of the final dataset sizes that are used in this work.

\begin{table}[]
    \centering
    \begin{tabular}{lrrr}
        \toprule
        \textbf{Dataset} & \textbf{Training} & \textbf{Validation} & \textbf{Testing} \\
        \toprule
        \multicolumn{4}{l}{\textit{Named-Entity Recognition (NER)}} \\
        Arabic & 3,973 & - & - \\
        Maltese & 155 & 43 & 2,109 \\
        \midrule
        \multicolumn{4}{l}{\textit{Sentiment Analysis (SA)}} \\
        Arabic & 15,305 & - & - \\
        Maltese & 595 & 85 & 171 \\
        \bottomrule
    \end{tabular}
    \caption{Data sizes for the downstream tasks}
    \label{table:data_sizes}
\end{table}

\subsection{Models and Inputs}
\label{section:models_inputs}

We experiment with the following models:
\begin{itemize}\itemsep0em
    \item\textbf{BERTu} \cite{micallef-etal-2022-pre}: a monolingual model pre-trained on Maltese.
    \item\textbf{mBERT} \cite{devlin-etal-2019-bert}: a multilingual model that includes Arabic in its pre-training, but not Maltese.
    \item\textbf{mBERTu} \cite{micallef-etal-2022-pre}: mBERT further pre-trained on Maltese.
\end{itemize}

In terms of Arabic inputs used for fine-tuning, we compare the original Arabic data in its original Arabic script (\textbf{Original}) with the different transliteration systems.
In addition to our transliteration systems -- \textbf{CharTx} and \textbf{MorphTx} -- we compare against two generic transliteration systems: \textbf{Buckwalter} and \textbf{Uroman}.
For Buckwalter we also lowercase the produced transliteration, since it uses uppercase letters for some of its mappings, while uppercase letters carry linguistic meaning in Latin-script languages.

We also include machine translation (\textbf{MT}) to measure the differences between translating and transliterating Arabic data.
We do so using Google Translate, from Arabic into Maltese.

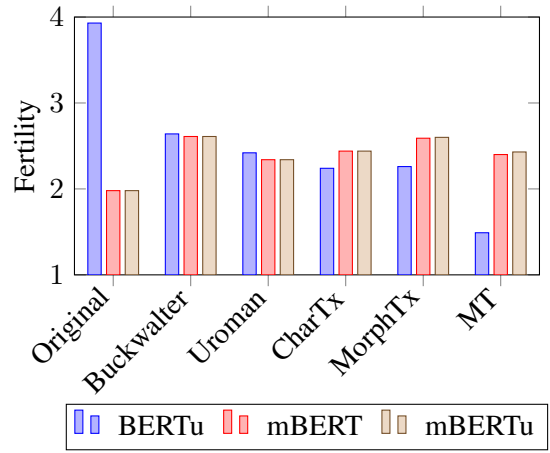
\begin{figure}[t]
    \centering

    \begin{tikzpicture}
        \begin{axis}[
        	ybar,
            ymin=1, ymax=4,
        	ylabel=Fertility,
            symbolic x coords={Original, Buckwalter, Uroman, CharTx, MorphTx, MT},
            xtick=data,
            bar width=5,
        	enlarge x limits=0.1,
            x tick label style={
                rotate=45,
                anchor=east,
            },
            y label style={
                at={(0.1, 0.5)},
            },
        	legend style={
                at={(0.5,-0.5)},
                anchor=north,
                legend columns=-1,
                align=center,
                column sep=5pt,
            },
            width=\linewidth,
            height=5cm,
        ]
            \addplot coordinates {
                (Original, 3.93)
                (Buckwalter, 2.64)
                (Uroman, 2.42)
                (CharTx, 2.24)
                (MorphTx, 2.26)
                (MT, 1.49)
            };
            \addlegendentry{BERTu}
            
            \addplot coordinates {
                (Original, 1.98)
                (Buckwalter, 2.61)
                (Uroman, 2.34)
                (CharTx, 2.44)
                (MorphTx, 2.59)
                (MT, 2.40)
            };
            \addlegendentry{mBERT}
            
            \addplot coordinates {
                (Original, 1.98)
                (Buckwalter, 2.61)
                (Uroman, 2.34)
                (CharTx, 2.44)
                (MorphTx, 2.60)
                (MT, 2.43)
            };
            \addlegendentry{mBERTu}
        \end{axis}
    \end{tikzpicture}
    
    \caption{Tokeniser fertility across datasets using the different Arabic inputs and models.}
    \label{figure:fertility}
\end{figure}

To analyse the impact of using different Arabic inputs, we compute the tokeniser fertility, which measures the average number of sub-tokens that the tokeniser splits a given token into \cite{acs-2019-vocabulary}.
For each input, we compute the fertility when processed with a model's tokeniser and visualise this in Figure~\ref{figure:fertility}.

BERTu with Original Arabic data has a high fertility (close to 4), but significantly drops with any of the transliteration inputs, dropping even further with machine translation.
We note that both our transliteration systems -- CharTx and MorphTx -- give lower fertility scores than the other transliteration systems, as we move closer towards Maltese orthography.
Fertility dropps even further with machine translated data.

Conversely, for mBERT and mBERTu, the lowest fertility is obtained with Original and increases with transliteration and machine translation, reflecting the overall lack of Maltese pre-training and vocabulary representation with these models.

\section{Transliteration Fine-Tuning Results}
\label{section:transliteration_results}

We first compare the different transliteration systems extrinsically with models fine-tuned on original Arabic as well as fine-tuning only on Maltese data.
Results are shown in Table~\ref{table:results_transliteration}.

\begin{table*}[t]
    \centering
    \tabcolsep3pt
    \begin{subtable}[t]{0.49\linewidth}
        \centering
        \begin{tabular}{lccc}
            \toprule
             & \textbf{BERTu} & \textbf{mBERT} & \textbf{mBERTu} \\
            \toprule
            \multicolumn{4}{p{0.9\columnwidth}}{\textit{Maltese data only}} \\
                               & 58.3 & 50.1 & 64.9 \\
            \midrule
            \multicolumn{4}{p{0.9\columnwidth}}{\textit{Adding Original Arabic data}} \\
                               & 55.7 & \textbf{60.8} & \textbf{69.5} \\
            \midrule
            \multicolumn{4}{p{0.9\columnwidth}}{\textit{Adding Transliterated Arabic data}} \\
            \textbf{Buckwalter}         & 55.9 & 51.4 & 63.7 \\
            \textbf{Uroman}             & 56.5 & 53.2 & 64.8 \\
            \textbf{CharTx}     & 59.3 & 54.3 & 65.6 \\
            \textbf{MorphTx}  & \textbf{61.3} & 55.5 & 67.0 \\
            \bottomrule
        \end{tabular}
        \caption{Named-Entity Recognition (NER)}
        \label{table:results_transliteration_ner}
    \end{subtable}
    \begin{subtable}[t]{0.49\linewidth}
        \centering
        \begin{tabular}{lccc}
            \toprule
             & \textbf{BERTu} & \textbf{mBERT} & \textbf{mBERTu} \\
            \toprule
            \multicolumn{4}{p{0.9\columnwidth}}{\textit{Maltese data only}} \\
                               & 82.5 & 65.4 & 80.7 \\
            \midrule
            \multicolumn{4}{p{0.9\columnwidth}}{\textit{Adding Original Arabic data}} \\
                               & 82.3 & \textbf{68.9} & 79.4 \\
            \midrule
            \multicolumn{4}{p{0.9\columnwidth}}{\textit{Adding Transliterated Arabic data}} \\
            \textbf{Buckwalter}         & 80.8 & 65.6 & 77.2 \\
            \textbf{Uroman}             & 83.6 & 67.3 & \textbf{81.4} \\
            \textbf{CharTx}     & \textbf{85.7} & 67.0 & 78.8 \\
            \textbf{MorphTx}  & 82.4 & 65.1 & 78.0 \\
            \bottomrule
        \end{tabular}
        \caption{Sentiment Analysis (SA)}
        \label{table:results_transliteration_ner}
    \end{subtable}

    \caption{
        Results from fine-tuning with transliterated Arabic, Arabic in the original script, and no Arabic data.
        The metrics used for NER and SA are span-level F1 and macro-averaged F1, respectively, and all scores are averages of 5 runs with different random seeds.
        Best scores per task and model are \textbf{bolded}.
    }
    \label{table:results_transliteration}
\end{table*}

\paragraph{No Arabic vs Original Arabic}
When adding original Arabic data, BERTu shows performance drops as expected, since it is only pre-trained on Maltese. However, mBERT and mBERTu generally improve with Arabic data, though mBERTu slightly underperforms on SA.

\paragraph{No Arabic vs Transliterated Arabic}
Adding transliterated data shows mixed results depending on the model and transliteration system. Buckwalter and Uroman underperform on NER for BERTu and mBERTu, though Uroman helps on SA. Conversely, mBERT benefits from both Buckwalter and Uroman. Our transliteration systems improve BERTu on both NER and SA tasks and outperforms Buckwalter for mBERT and mBERTu, with CharTx best on NER and MorphTx best on SA.

\paragraph{Original Arabic vs Transliterated Arabic}
Comparing the performance of adding Arabic data in different scripts we also observe similar trends.
Except for Buckwalter, BERTu consistently attains better performance with transliterated data.
In contrast, both mBERT and mBERTu perform worse with transliterated data, except for Uroman on SA.

In summary, data augmentation with some form of Arabic data generally improves performance.
However, performance improvements with transliteration is dependent on the model's exposure to Maltese.
Crucially, we highlight the importance of applying an appropriate transliteration scheme -- while transliteration can eliminate script differences, its effectiveness relies on the orthographic similarities with the target language.

\section{Cascaded Arabic Training}
\label{section:cascaded_results}

In Section~\ref{section:transliteration_results}, our results show that multilingual models do not benefit from cross-lingual transfer capabilities as much when fine-tuning with transliterated Arabic, instead of original Arabic.
We hypothesise that this is due the model's pre-training on Arabic in Arabic script rather than Latin.

To test this, we conduct further experiments where we cascade multiple stages of fine-tuning on Arabic, with increasing similarity to Maltese, before the final stage of Maltese fine-tuning.
Therefore, we first start with a fine-tuning step on original Arabic data, followed by transliterated Arabic data, and lastly a final phase of Maltese fine-tuning.
To simplify the setup, we choose one transliteration system based on observed performance trends from Section~\ref{section:transliteration_results} -- MorphTx for NER and CharTx for SA.
Furthermore, we consider another variant where we also fine-tune on machine translated Arabic after fine-tuning on original and transliterated Arabic.

We compare these cascaded approaches against fine-tuning with only one stage of Arabic fine-tuning -- original, transliterated, and machine translated Arabic.
The results are shown in Table~\ref{table:results_cascade}.\footnote{
The numbers from Table~\ref{table:results_transliteration} for Maltese only, original Arabic, and transliterated Arabic, are relisted to ease comparison.
}

\begin{table*}[t]
    \centering
    \tabcolsep3pt
    \begin{subtable}[t]{0.49\linewidth}
        \centering
        \begin{tabular}{ccc ccc}
            \toprule
            \textbf{Orig} & \textbf{Tx} &\textbf{MT} & \textbf{BERTu} & \textbf{mBERT} & \textbf{mBERTu} \\
            \toprule
            \multicolumn{6}{p{0.9\columnwidth}}{\textit{Maltese data only}} \\
                       &            &            & 58.3 & 50.1 & 64.9 \\
            \midrule
            \multicolumn{6}{p{0.9\columnwidth}}{\textit{Adding a single stage of Arabic fine-tuning}} \\
            \checkmark &            &            & 55.7 & 60.8 & 69.5 \\
                       & \checkmark &            & 61.3 & 55.5 & 67.0 \\
                       &            & \checkmark & \textbf{67.3} & 59.4 & 69.1 \\
            \midrule
            \multicolumn{6}{p{0.9\columnwidth}}{\textit{Adding multiple stages of Arabic fine-tuning}} \\
            \checkmark & \checkmark &            & 58.6 & 61.6 & \textbf{70.2} \\
            \checkmark & \checkmark & \checkmark & 61.8 & \textbf{61.8} & 68.8 \\
            \bottomrule
        \end{tabular}
        \caption{Named-Entity Recognition (NER)}
        \label{results_cascade_ner}
    \end{subtable}
    \begin{subtable}[t]{0.49\linewidth}
        \centering
        \begin{tabular}{ccc ccc}
            \toprule
            \textbf{Orig} & \textbf{Tx} &\textbf{MT} & \textbf{BERTu} & \textbf{mBERT} & \textbf{mBERTu} \\
            \toprule
            \multicolumn{6}{p{0.9\columnwidth}}{\textit{Maltese data only}} \\
                       &            &            & 82.5 & 65.4 & \textbf{80.7} \\
            \midrule
            \multicolumn{6}{p{0.9\columnwidth}}{\textit{Adding a single stage of Arabic fine-tuning}} \\
            \checkmark &            &            & 82.3 & 68.9 & 79.4 \\
                       & \checkmark &            & \textbf{85.7} & 67.0 & 78.8 \\
                       &            & \checkmark & 82.5 & 69.3 & 78.7 \\
            \midrule
            \multicolumn{6}{p{0.9\columnwidth}}{\textit{Adding multiple stages of Arabic fine-tuning}} \\
            \checkmark & \checkmark &            & 84.0 & 69.3 & 79.8 \\
            \checkmark & \checkmark & \checkmark & 82.7 & \textbf{70.0} & 76.7 \\
            \bottomrule
        \end{tabular}
        \caption{Sentiment Analysis (SA)}
        \label{results_cascade_ner}
    \end{subtable}
    
    \caption{
        Results from fine-tuning with different Arabic inputs: Original (Orig), Transliteration (Tx -- MorphTx for NER and CharTx for SA), and Machine Translation (MT).
        The metrics used for NER and SA are span-level F1 and macro-averaged F1, respectively, and all scores are averages of 5 runs with different random seeds.
        Best scores per task and model are \textbf{bolded}.
    }
    \label{table:results_cascade}
\end{table*}

From our results, we observe that fine-tuning with MT is a competitive baseline.
When fine-tuning with translations, better performance is obtained than all the previously presented approaches on NER with BERTu and on SA with mBERT.

With BERTu, cascaded Arabic fine-tuning does not help over a single phase of Arabic fine-tuning, as the best result is obtained with machine translation for NER and transliteration for SA.
This is likely due BERTu's Maltese pre-training with little to no Arabic data.
This is supported by the relatively lower scores observed for BERTu when fine-tuned with original Arabic.

In contrast, both mBERT and mBERTu achieve the best performance when fine-tuned on Arabic data in a cascaded approach.
For mBERTu, the best result is obtained with original and transliteration cascading, whereas the full cascade yields the best result for mBERT.
Interestingly, when mBERT and mBERTu are fine-tuned solely with either transliterated Arabic or translated Arabic, worse results are generally obtained compared to original Arabic fine-tuning, but by combining all approaches in one pipeline, we are better able to unlock the models' cross-lingual transfer capabilities.

\section{Conclusion and Future Work}

In this work, we presented transliteration systems from Arabic to Maltese.
Our experimental results highlight that the effectiveness of transliteration depends on the model's exposure to the target language.
We find that a monolingual Maltese model benefits from transliterated Arabic data, while multilingual models are able to make cross-lingual transfer links without transliteration.
However, cascading the original and transliterated data during the fine-tuning process proves to be beneficial over fine-tuning with only one of these.

Future work includes exploring unsupervised or self-supervised methods to better align Arabic and Maltese representations without heavy reliance on parallel data or linguistic analysis.
We also plan to investigate advanced machine translation and neural transliteration models that capture deeper morphological and phonological patterns.
Additionally, we plan on applying these augmentation techniques to other facets of Maltese NLP such as language modelling.

\section{Limitations}
\label{section:limitations}

While our results demonstrate the potential of Arabic-driven augmentation for Maltese NLP, several limitations remain. First, the effectiveness of our approach is partly constrained by the quality of Arabic-to-Maltese translations, which may introduce stylistic or grammatical inconsistencies, especially when using off-the-shelf machine translation systems. Second, our transliteration rules and mappings, though linguistically motivated, simplify complex morphological and phonological relationships and may not generalise across all domains of Maltese usage. Third, our evaluations focus on downstream tasks using pre-existing datasets, which may not fully capture real-world variation in code-switching, informal registers, or dialectal usage. Finally, while we focus on Standard Maltese and Modern Standard Arabic, variation across dialects and registers in both languages is not addressed and may affect generalisability.

\section{Ethical Considerations}
\label{section:ethics}

Our study uses publicly available language resources and models for both Arabic and Maltese, adhering to the licencing and usage terms of each dataset. However, the use of machine translation for data augmentation carries the risk of reinforcing biases or introducing artefacts that may impact fairness and interpretability in downstream tasks. Moreover, while our approach aims to support a low-resource language, it assumes a certain level of equivalence between Arabic and Maltese that may obscure sociolinguistic or cultural distinctions. We encourage careful application of our methods, particularly in contexts involving sensitive or identity-related content. We also note that our transliteration and augmentation techniques are not intended for human communication and may not reflect idiomatic or culturally appropriate usage.

We used AI writing assistance within the scope of ``Assistance purely with the language of the paper'' described in the ACL Policy on Publication Ethics.

\section*{Acknowledgments}

We thank Salam Khalifa and Rawan Bondok for their helpful feedback when preparing this paper.
We also acknowledge funding by Malta Enterprise Research and Development Scheme.

\bibliography{anthology,custom}

\begin{thebibliography}{26}
\providecommand{\natexlab}[1]{#1}

\bibitem[{{\'{A}}cs(2019)}]{acs-2019-vocabulary}
Judit {\'{A}}cs. 2019.
\newblock \href {https://juditacs.github.io/2019/02/19/bert-tokenization-stats.html} {Exploring {BERT}'s vocabulary}.
\newblock \textit{Judit {\'{A}}cs's blog} (Accessed 2025-05-15).

\bibitem[{Baly et~al.(2018)Baly, Khaddaj, Hajj, El-Hajj, and Shaban}]{baly-etal-2018-arsentdlev}
Ramy Baly, Alaa Khaddaj, Hazem Hajj, Wassim El-Hajj, and Khaled Shaban. 2018.
\newblock \href {http://lrec-conf.org/workshops/lrec2018/W30/summaries/23_W30.html} {Arsentd-lev: A multi-topic corpus for target-based sentiment analysis in arabic levantine tweets}.
\newblock In \emph{Proceedings of the Eleventh International Conference on Language Resources and Evaluation (LREC 2018)}, Paris, France. European Language Resources Association (ELRA).

\bibitem[{Benajiba et~al.(2007)Benajiba, Rosso, and Bened{\'i}~Ruiz}]{benajiba-etal-2007-anersys}
Yassine Benajiba, Paolo Rosso, and Jos{\'e}~Miguel Bened{\'i}~Ruiz. 2007.
\newblock \href {https://doi.org/10.1007/978-3-540-70939-8_13} {{ANERsys}: An arabic named entity recognition system based on maximum entropy}.
\newblock In \emph{Computational Linguistics and Intelligent Text Processing}, pages 143--153, Berlin, Heidelberg. Springer Berlin Heidelberg.

\bibitem[{Buckwalter(2002)}]{Buckwalter:2002:buckwalter}
Tim Buckwalter. 2002.
\newblock Buckwalter {{A}rabic} morphological analyzer version 1.0.
\newblock Linguistic Data Consortium (LDC) catalog number LDC2002L49, ISBN 1-58563-257-0.

\bibitem[{Devlin et~al.(2019)Devlin, Chang, Lee, and Toutanova}]{devlin-etal-2019-bert}
Jacob Devlin, Ming-Wei Chang, Kenton Lee, and Kristina Toutanova. 2019.
\newblock \href {https://doi.org/10.18653/v1/N19-1423} {{BERT}: Pre-training of deep bidirectional transformers for language understanding}.
\newblock In \emph{Proceedings of the 2019 Conference of the North {A}merican Chapter of the Association for Computational Linguistics: Human Language Technologies, Volume 1 (Long and Short Papers)}, pages 4171--4186, Minneapolis, Minnesota. Association for Computational Linguistics.

\bibitem[{Eryani and Habash(2021)}]{eryani-habash-2021-automatic}
Fadhl Eryani and Nizar Habash. 2021.
\newblock \href {https://aclanthology.org/2021.wanlp-1.23/} {Automatic {R}omanization of {A}rabic bibliographic records}.
\newblock In \emph{Proceedings of the Sixth Arabic Natural Language Processing Workshop}, pages 213--218, Kyiv, Ukraine (Virtual). Association for Computational Linguistics.

\bibitem[{Gianola et~al.(2020)Gianola, Ēriks Ajausks, Arranz, Bendahman, Bié, Borg, Cerdà, Choukri, Cuadros, de~Gibert, Degroote, Edelman, Etchegoyhen, Ángela Franco~Torres, Hernandez, Pablos, Gatt, Grouin, Herranz, Kohan, Lavergne, Melero, Paroubek, Rigault, Rosner, Rozis, van~der Plas, Vīksna, and Zweigenbaum}]{gianola-etal-2020-mapa}
Lucie Gianola, Ēriks Ajausks, Victoria Arranz, Chomicha Bendahman, Laurent Bié, Claudia Borg, Aleix Cerdà, Khalid Choukri, Montse Cuadros, Ona de~Gibert, Hans Degroote, Elena Edelman, Thierry Etchegoyhen, Ángela Franco~Torres, Mercedes~García Hernandez, Aitor~García Pablos, Albert Gatt, Cyril Grouin, Manuel Herranz, and 10 others. 2020.
\newblock \href {https://doi.org/10.3233/FAIA200869} {Automatic removal of identifying information in official eu languages for public administrations: The {MAPA} project}.
\newblock In \emph{Proceedings of the 33rd International Conference on Legal Knowledge and Information Systems ({JURIX'20})}, pages 223--226. IOS Press.

\bibitem[{Habash et~al.(2018)Habash, Eryani, Khalifa, Rambow, Abdulrahim, Erdmann, Faraj, Zaghouani, Bouamor, Zalmout, Hassan, Al-Shargi, Alkhereyf, Abdulkareem, Eskander, Salameh, and Saddiki}]{habash-etal-2018-unified}
Nizar Habash, Fadhl Eryani, Salam Khalifa, Owen Rambow, Dana Abdulrahim, Alexander Erdmann, Reem Faraj, Wajdi Zaghouani, Houda Bouamor, Nasser Zalmout, Sara Hassan, Faisal Al-Shargi, Sakhar Alkhereyf, Basma Abdulkareem, Ramy Eskander, Mohammad Salameh, and Hind Saddiki. 2018.
\newblock \href {https://aclanthology.org/L18-1574/} {Unified guidelines and resources for {A}rabic dialect orthography}.
\newblock In \emph{Proceedings of the Eleventh International Conference on Language Resources and Evaluation ({LREC} 2018)}, Miyazaki, Japan. European Language Resources Association (ELRA).

\bibitem[{Habash et~al.(2012)Habash, Eskander, and Hawwari}]{habash-etal-2012-morphological}
Nizar Habash, Ramy Eskander, and Abdelati Hawwari. 2012.
\newblock \href {https://aclanthology.org/W12-2301/} {A morphological analyzer for {E}gyptian {A}rabic}.
\newblock In \emph{Proceedings of the Twelfth Meeting of the Special Interest Group on Computational Morphology and Phonology}, pages 1--9, Montr{\'e}al, Canada. Association for Computational Linguistics.

\bibitem[{Habash(2010)}]{habash-2010-arabic-nlp}
Nizar~Y Habash. 2010.
\newblock \href {https://doi.org/10.1007/978-3-031-02139-8} {\emph{Introduction to {A}rabic Natural Language Processing}}.
\newblock Morgan \& Claypool Publishers.

\bibitem[{Hermjakob et~al.(2018)Hermjakob, May, and Knight}]{hermjakob-etal-2018-box}
Ulf Hermjakob, Jonathan May, and Kevin Knight. 2018.
\newblock \href {https://doi.org/10.18653/v1/P18-4003} {Out-of-the-box universal {R}omanization tool uroman}.
\newblock In \emph{Proceedings of {ACL} 2018, System Demonstrations}, pages 13--18, Melbourne, Australia. Association for Computational Linguistics.

\bibitem[{Inoue et~al.(2022)Inoue, Khalifa, and Habash}]{inoue-etal-2022-morphosyntactic}
Go~Inoue, Salam Khalifa, and Nizar Habash. 2022.
\newblock \href {https://doi.org/10.18653/v1/2022.findings-acl.135} {Morphosyntactic tagging with pre-trained language models for {A}rabic and its dialects}.
\newblock In \emph{Findings of the Association for Computational Linguistics: ACL 2022}, pages 1708--1719, Dublin, Ireland. Association for Computational Linguistics.

\bibitem[{Lauscher et~al.(2020)Lauscher, Ravishankar, Vuli{\'c}, and Glava{\v{s}}}]{lauscher-etal-2020-zero}
Anne Lauscher, Vinit Ravishankar, Ivan Vuli{\'c}, and Goran Glava{\v{s}}. 2020.
\newblock \href {https://doi.org/10.18653/v1/2020.emnlp-main.363} {From zero to hero: {O}n the limitations of zero-shot language transfer with multilingual {T}ransformers}.
\newblock In \emph{Proceedings of the 2020 Conference on Empirical Methods in Natural Language Processing (EMNLP)}, pages 4483--4499, Online. Association for Computational Linguistics.

\bibitem[{Liu et~al.(2025)Liu, Ma, Ye, and Sch{\"u}tze}]{liu-etal-2025-transmi}
Yihong Liu, Chunlan Ma, Haotian Ye, and Hinrich Sch{\"u}tze. 2025.
\newblock \href {https://aclanthology.org/2025.coling-main.32/} {{T}rans{MI}: A framework to create strong baselines from multilingual pretrained language models for transliterated data}.
\newblock In \emph{Proceedings of the 31st International Conference on Computational Linguistics}, pages 469--495, Abu Dhabi, UAE. Association for Computational Linguistics.

\bibitem[{Mart{\'i}nez-Garc{\'i}a et~al.(2021)Mart{\'i}nez-Garc{\'i}a, Badia, and Barnes}]{martinez-garcia-etal-2021-evaluating}
Antonio Mart{\'i}nez-Garc{\'i}a, Toni Badia, and Jeremy Barnes. 2021.
\newblock \href {https://doi.org/10.18653/v1/2021.acl-long.244} {Evaluating morphological typology in zero-shot cross-lingual transfer}.
\newblock In \emph{Proceedings of the 59th Annual Meeting of the Association for Computational Linguistics and the 11th International Joint Conference on Natural Language Processing (Volume 1: Long Papers)}, pages 3136--3153, Online. Association for Computational Linguistics.

\bibitem[{Micallef et~al.(2023)Micallef, Eryani, Habash, Bouamor, and Borg}]{micallef-etal-2023-exploring}
Kurt Micallef, Fadhl Eryani, Nizar Habash, Houda Bouamor, and Claudia Borg. 2023.
\newblock \href {https://doi.org/10.18653/v1/2023.cawl-1.4} {Exploring the impact of transliteration on {NLP} performance: Treating {M}altese as an {A}rabic dialect}.
\newblock In \emph{Proceedings of the Workshop on Computation and Written Language (CAWL 2023)}, pages 22--32, Toronto, Canada. Association for Computational Linguistics.

\bibitem[{Micallef et~al.(2022)Micallef, Gatt, Tanti, van~der Plas, and Borg}]{micallef-etal-2022-pre}
Kurt Micallef, Albert Gatt, Marc Tanti, Lonneke van~der Plas, and Claudia Borg. 2022.
\newblock \href {https://doi.org/10.18653/v1/2022.deeplo-1.10} {Pre-training data quality and quantity for a low-resource language: New corpus and {BERT} models for {M}altese}.
\newblock In \emph{Proceedings of the Third Workshop on Deep Learning for Low-Resource Natural Language Processing}, pages 90--101, Hybrid. Association for Computational Linguistics.

\bibitem[{Micallef et~al.(2024)Micallef, Habash, Borg, Eryani, and Bouamor}]{micallef-etal-2024-cross}
Kurt Micallef, Nizar Habash, Claudia Borg, Fadhl Eryani, and Houda Bouamor. 2024.
\newblock \href {https://aclanthology.org/2024.eacl-long.61/} {Cross-lingual transfer from related languages: Treating low-resource {M}altese as multilingual code-switching}.
\newblock In \emph{Proceedings of the 18th Conference of the European Chapter of the Association for Computational Linguistics (Volume 1: Long Papers)}, pages 1014--1025, St. Julian{'}s, Malta. Association for Computational Linguistics.

\bibitem[{Muller et~al.(2021)Muller, Anastasopoulos, Sagot, and Seddah}]{muller-etal-2021-unseen}
Benjamin Muller, Antonios Anastasopoulos, Beno{\^i}t Sagot, and Djam{\'e} Seddah. 2021.
\newblock \href {https://doi.org/10.18653/v1/2021.naacl-main.38} {When being unseen from m{BERT} is just the beginning: Handling new languages with multilingual language models}.
\newblock In \emph{Proceedings of the 2021 Conference of the North American Chapter of the Association for Computational Linguistics: Human Language Technologies}, pages 448--462, Online. Association for Computational Linguistics.

\bibitem[{Obeid et~al.(2020)Obeid, Zalmout, Khalifa, Taji, Oudah, Alhafni, Inoue, Eryani, Erdmann, and Habash}]{obeid-etal-2020-camel}
Ossama Obeid, Nasser Zalmout, Salam Khalifa, Dima Taji, Mai Oudah, Bashar Alhafni, Go~Inoue, Fadhl Eryani, Alexander Erdmann, and Nizar Habash. 2020.
\newblock \href {https://aclanthology.org/2020.lrec-1.868/} {{CAM}e{L} tools: An open source python toolkit for {A}rabic natural language processing}.
\newblock In \emph{Proceedings of the Twelfth Language Resources and Evaluation Conference}, pages 7022--7032, Marseille, France. European Language Resources Association.

\bibitem[{Schmidt et~al.(2022)Schmidt, Vuli{\'c}, and Glava{\v{s}}}]{schmidt-etal-2022-dont}
Fabian~David Schmidt, Ivan Vuli{\'c}, and Goran Glava{\v{s}}. 2022.
\newblock \href {https://doi.org/10.18653/v1/2022.emnlp-main.736} {Don`t stop fine-tuning: On training regimes for few-shot cross-lingual transfer with multilingual language models}.
\newblock In \emph{Proceedings of the 2022 Conference on Empirical Methods in Natural Language Processing}, pages 10725--10742, Abu Dhabi, United Arab Emirates. Association for Computational Linguistics.

\bibitem[{Winata et~al.(2022)Winata, Wu, Kulkarni, Solorio, and Preotiuc-Pietro}]{winata-etal-2022-cross}
Genta Winata, Shijie Wu, Mayank Kulkarni, Thamar Solorio, and Daniel Preotiuc-Pietro. 2022.
\newblock \href {https://doi.org/10.18653/v1/2022.aacl-main.59} {Cross-lingual few-shot learning on unseen languages}.
\newblock In \emph{Proceedings of the 2nd Conference of the Asia-Pacific Chapter of the Association for Computational Linguistics and the 12th International Joint Conference on Natural Language Processing (Volume 1: Long Papers)}, pages 777--791, Online only. Association for Computational Linguistics.

\bibitem[{Wolf et~al.(2020)Wolf, Debut, Sanh, Chaumond, Delangue, Moi, Cistac, Rault, Louf, Funtowicz, Davison, Shleifer, von Platen, Ma, Jernite, Plu, Xu, Le~Scao, Gugger, Drame, Lhoest, and Rush}]{wolf-etal-2020-transformers}
Thomas Wolf, Lysandre Debut, Victor Sanh, Julien Chaumond, Clement Delangue, Anthony Moi, Pierric Cistac, Tim Rault, Remi Louf, Morgan Funtowicz, Joe Davison, Sam Shleifer, Patrick von Platen, Clara Ma, Yacine Jernite, Julien Plu, Canwen Xu, Teven Le~Scao, Sylvain Gugger, and 3 others. 2020.
\newblock \href {https://doi.org/10.18653/v1/2020.emnlp-demos.6} {Transformers: State-of-the-art natural language processing}.
\newblock In \emph{Proceedings of the 2020 Conference on Empirical Methods in Natural Language Processing: System Demonstrations}, pages 38--45, Online. Association for Computational Linguistics.

\bibitem[{Wu and Dredze(2019)}]{wu-dredze-2019-beto}
Shijie Wu and Mark Dredze. 2019.
\newblock \href {https://doi.org/10.18653/v1/D19-1077} {Beto, bentz, becas: The surprising cross-lingual effectiveness of {BERT}}.
\newblock In \emph{Proceedings of the 2019 Conference on Empirical Methods in Natural Language Processing and the 9th International Joint Conference on Natural Language Processing (EMNLP-IJCNLP)}, pages 833--844, Hong Kong, China. Association for Computational Linguistics.

\bibitem[{Wu and Dredze(2020)}]{wu-dredze-2020-languages}
Shijie Wu and Mark Dredze. 2020.
\newblock \href {https://doi.org/10.18653/v1/2020.repl4nlp-1.16} {Are all languages created equal in multilingual {BERT}?}
\newblock In \emph{Proceedings of the 5th Workshop on Representation Learning for NLP}, pages 120--130, Online. Association for Computational Linguistics.

\bibitem[{Zhao et~al.(2021)Zhao, Zhu, Shareghi, Vuli{\'c}, Reichart, Korhonen, and Sch{\"u}tze}]{zhao-etal-2021-closer}
Mengjie Zhao, Yi~Zhu, Ehsan Shareghi, Ivan Vuli{\'c}, Roi Reichart, Anna Korhonen, and Hinrich Sch{\"u}tze. 2021.
\newblock \href {https://doi.org/10.18653/v1/2021.acl-long.447} {A closer look at few-shot crosslingual transfer: The choice of shots matters}.
\newblock In \emph{Proceedings of the 59th Annual Meeting of the Association for Computational Linguistics and the 11th International Joint Conference on Natural Language Processing (Volume 1: Long Papers)}, pages 5751--5767, Online. Association for Computational Linguistics.

\end{thebibliography}

\appendix
\newpage

\section{Transliteration System Rules}
\label{appendix:rules}

The character mappings presented in Section~\ref{section:transliteration_character} are presented in Tables~\ref{table:character_mappings_multiple} and~\ref{table:character_mappings_single} while the morpheme mappings presented in Section~\ref{section:transliteration_morphology} are shown in Table~\ref{table:morpheme_mappings}.
We note how many of the character mappings include diacritics on the source side and are typically not used by the CharTx system (Section~\ref{section:transliteration_character}) since the text is not diacritised.

\begin{table*}
\centering
\begin{subtable}[t]{0.5\linewidth}
    \centering
    \begin{tabular}[t]{rc}
        \toprule
        \multicolumn{1}{c}{\textbf{Source}} & \multicolumn{1}{c}{\textbf{Target}} \\
        \toprule
        \multicolumn{2}{l}{\textbf{\<ل> \textit{l} + Sun Letter with Gemination}} \\
        \<لدّ> \textit{ld\textasciitilde} / \<لذّ> \textit{l*\textasciitilde} / \<لضّ> \textit{lD\textasciitilde} / \<لظّ> \textit{lZ\textasciitilde} & dd \\
        \<للّ> \textit{ll\textasciitilde} & ll \\
        \<لنّ> \textit{ln\textasciitilde} & nn \\
        \<لرّ> \textit{lr\textasciitilde} & rr \\
        \<لسّ> \textit{ls\textasciitilde} / \<لصّ> \textit{lS\textasciitilde} & ss \\
        \<لتّ> \textit{lt\textasciitilde} / \<لثّ> \textit{lv\textasciitilde} / \<لطّ> \textit{lT\textasciitilde} & tt \\
        \<لشّ> \textit{l\$\textasciitilde} & xx \\
        \<لزّ> \textit{lz\textasciitilde} & żż \\
        \midrule
        \multicolumn{2}{l}{\textbf{Final \<ي> \textit{y} with Gemination}} \\
        \texttt{[EOS]}\<يّ> \textit{y\textasciitilde} & i \\
        \midrule
        \multicolumn{2}{l}{\textbf{Gemination}} \\
        \<بّ> \textit{b\textasciitilde} & bb \\
        \<دّ> \textit{d\textasciitilde} / \<ذّ> \textit{*\textasciitilde} / \<ضّ> \textit{D\textasciitilde} / \<ظّ> \textit{Z\textasciitilde} & dd \\
        \<فّ> \textit{f\textasciitilde} & ff \\
        \<هّ> \textit{h\textasciitilde} & hh \\
        \<جّ> \textit{j\textasciitilde} & ġġ \\
        \<حّ> \textit{H\textasciitilde} / \<خّ> \textit{x\textasciitilde} & \mh\mh \\
        \<عّ> \textit{E\textasciitilde} / \<غّ> \textit{g\textasciitilde} & g\mh \\
        \<يّ> \textit{y\textasciitilde} & jj \\
        \<كّ> \textit{k\textasciitilde} & kk \\
        \<لّ> \textit{l\textasciitilde} & ll \\
        \<مّ> \textit{m\textasciitilde} & mm \\
        \<نّ> \textit{n\textasciitilde} & nn \\
        \<قّ> \textit{q\textasciitilde} & qq \\
        \<رّ> \textit{r\textasciitilde} & rr \\
        \<سّ> \textit{s\textasciitilde} / \<صّ> \textit{S\textasciitilde} & ss \\
        \<تّ> \textit{t\textasciitilde} / \<ثّ> \textit{v\textasciitilde} / \<طّ> \textit{T\textasciitilde} & tt \\
        \<وّ> \textit{w\textasciitilde} & ww \\
        \<شّ> \textit{\$\textasciitilde} & xx \\
        \<زّ> \textit{z\textasciitilde} & żż \\
        \bottomrule
    \end{tabular}
\end{subtable}
\begin{subtable}[t]{0.45\linewidth}
    \centering
    \begin{tabular}[t]{rc}
        \toprule
        \multicolumn{1}{c}{\textbf{Source}} & \multicolumn{1}{c}{\textbf{Target}} \\
        \toprule
        \multicolumn{2}{l}{\textbf{Hamza/Alif with Diacritic}} \\
        \<ءْ> \textit{'o} / \<اً> \textit{AF} / \<اٍ> \textit{AK} / \<اٌ> \textit{AN} &  \\
        \<ءَ> \textit{'a} / \<أَ> \textit{>a} / \alifwaslafatha{} \textit{\{a} / \<آَ> \textit{|a} & a \\
        \<ءِ> \textit{'i} / \<إِ> \textit{<i} / \alifwaslakasra{} \textit{\{i} / \<آِ> \textit{|i} & i \\
        \<ءُ> \textit{'u} / \<أُ> \textit{>u} / \alifwasladamma{} \textit{\{u} / \<آُ> \textit{|u} & u \\
        \midrule
        \multicolumn{2}{l}{\textbf{Long Vowel `\textit{a}'}} \\
        \<ا>\fatha{} \textit{aA} & a \\
        \<ى>\fatha{} \textit{aY} & a \\
        \<ة>\fatha{} \textit{ap} & a \\
        \midrule
        \multicolumn{2}{l}{\textbf{`\textit{i}'/`\textit{y}' Glide}} \\
        \<يا> \textit{yA} & ja \\
        \<يْ> \textit{yo} & j \\
        \<يَ> \textit{ya} / \<ئَ> \textit{\}a} & je \\
        \<يِ> \textit{yi} / \<ئِ> \textit{\}i} & ji \\
        \<يُ> \textit{yu} / \<ئُ> \textit{\}u} & ju \\
        \<ي>\skun{} \textit{oy} & j \\
        \<ي>\fatha{} \textit{ay} & ej \\
        \<ي>\damma{} \textit{uy} & uj \\
        \<ي>\kasra{} \textit{iy} & i \\
        \<ئْ> \textit{\}o} & i \\
        \midrule
        \multicolumn{2}{l}{\textbf{`\textit{u}'/`\textit{w}' Glide}} \\
        \<وا> \textit{wA} & wa \\
        \<وْ> \textit{wo} & w \\
        \<وَ> \textit{wa} / \<ؤَ> \textit{\&a} & we \\
        \<وِ> \textit{wi} / \<ؤِ> \textit{\&i} & wi \\
        \<وُ> \textit{wu} / \<ؤُ> \textit{\&u} & wu \\
        \<و>\skun{} \textit{ow} & w \\
        \<و>\fatha{} \textit{aw} & ew \\
        \<و>\kasra{} \textit{iw} & iw \\
        \<و>\damma{} \textit{uw} & u \\
        \<ؤْ> \textit{\&o} & u \\
        \bottomrule
    \end{tabular}
\end{subtable}
\caption{
    Character Mappings (multiple characters) outlining how source Arabic characters are mapped to target Maltese characters.
    The Buckwalter representation for Arabic is also shown.
    \texttt{[BOS]} and \texttt{[EOS]} are special markers indicating the beginning and end positions of a word.
}
\label{table:character_mappings_multiple}
\end{table*}

\begin{table*}
\centering
\begin{subtable}[t]{0.6\linewidth}
    \centering
    \begin{tabular}[t]{rc}
        \toprule
        \multicolumn{1}{c}{\textbf{Source}} & \multicolumn{1}{c}{\textbf{Target}} \\
        \toprule
        \multicolumn{2}{l}{\textbf{Diacritics}} \\
        \alif{} \textit{\textasciigrave} & a \\
        \fatha{} \textit{a} & e \\
        \kasra{} \textit{i} & i \\
        \damma{} \textit{u} & u \\
        \skun{} \textit{o} / \fathatan{} \textit{F} / \kasratan{} \textit{K} / \dammatan{} \textit{N} / \shaddah{} \textit{\textasciitilde} &  \\
        \midrule
        \multicolumn{2}{l}{\textbf{Special Characters at Word Boundaries}} \\
        \texttt{[EOS]}\<ع> \textit{E} / \texttt{[EOS]}\<غ> \textit{g} & ' \\
        \<أ>\texttt{[BOS]} \textit{>} / \<إ>\texttt{[BOS]} \textit{<} &  \\
        \midrule
        \multicolumn{2}{l}{\textbf{Letters}} \\
        \<ا> \textit{A} / \<أ> \textit{>} / \<آ> \textit{|} / \<ى> \textit{Y} & a \\
        \<ي> \textit{y} / \<إ> \textit{<} / \<ئ> \textit{\}} & i \\
        \<و> \textit{w} / \<ؤ> \textit{\&} & u \\
        \<ء> \textit{'} / \<ٱ> \textit{\{} &  \\
        \<ة> \textit{p} & a \\
        \<ب> \textit{b} & b \\
        \<د> \textit{d} / \<ذ> \textit{*} / \<ض> \textit{D} / \<ظ> \textit{Z} & d \\
        \<ف> \textit{f} & f \\
        \<ج> \textit{j} & ġ \\
        \<ه> \textit{h} & h \\
        \<ح> \textit{H} / \<خ> \textit{x} & \mh \\
        \<ع> \textit{E} / \<غ> \textit{g} & g\mh \\
        \<ك> \textit{k} & k \\
        \<ل> \textit{l} & l \\
        \<م> \textit{m} & m \\
        \<ن> \textit{n} & n \\
        \<ق> \textit{q} & q \\
        \<ر> \textit{r} & r \\
        \<س> \textit{s} / \<ص> \textit{S} & s \\
        \<ت> \textit{t} / \<ث> \textit{v} / \<ط> \textit{T} & t \\
        \<ش> \textit{\$} & x \\
        \<ز> \textit{z} & ż \\
        \bottomrule
    \end{tabular}
\end{subtable}
\begin{subtable}[t]{0.35\linewidth}
    \centering
    \begin{tabular}[t]{rc}
        \toprule
        \multicolumn{1}{c}{\textbf{Source}} & \multicolumn{1}{c}{\textbf{Target}} \\
        \toprule
        \multicolumn{2}{l}{\textbf{Symbols}} \\
        \<،> & , \\
        \<؛> & ; \\
        \<؟> & ? \\
        \<٪> & \% \\
        \<٩> & 9 \\
        \<٨> & 8 \\
        \<٧> & 7 \\
        \<٦> & 6 \\
        \<٥> & 5 \\
        \<٤> & 4 \\
        \<٣> & 3 \\
        \<٢> & 2 \\
        \<١> & 1 \\
        \<٠> & 0 \\
        \bottomrule
    \end{tabular}
\end{subtable}
\caption{
    Character Mappings (single characters) outlining how source Arabic characters are mapped to target Maltese characters.
    The Buckwalter representation for Arabic is also shown.
    \texttt{[BOS]} and \texttt{[EOS]} are special markers indicating the beginning and end positions of a word.
}
\label{table:character_mappings_single}
\end{table*}

\begin{table*}[t]
    \centering
    \begin{tabular}{lrl}
        \toprule
        \textbf{Tag} & \textbf{Source} & \textbf{Target} \\
        \toprule
        CONJ & \<وَ> \textit{wa} / \<وِ> \textit{wi} & u\_ \\
        DET & \<ال> \textit{Al} & il- \\
        PREP & \<بِ> \textit{bi} & bi\_ \\
        PREP & \<لِ> \textit{li} & li\_ \\
        PREP & \<فِي> \textit{fiy} & fi\_ \\
        NOUN & \<مَع> \textit{maE} & ma'\_ \\
        NOUN & \<تَاع> \textit{taAE} & ta'\_ \\
        PREP & \<عَلَى> \textit{EalaY} & g\mh al\_ \\
        PREP & \<مِن> \textit{min} & minn\_ \\
        NSUFF\_FEM\_SG (construct state) & \<ة> \textit{p} & t \\
        FUT\_PART & \<سَ> \textit{sa} & sa \\
        CASE\_*\_* & * & \\
        IVSUFF\_MOOD:* & * & \\
        IVSUFF\_SUBJ:2FS & * & \\
        IVSUFF\_SUBJ:\{D,MP,FP\} & * & u \\
        PVSUFF\_SUBJ:\{1S,2MS,2FS\} & * & t \\
        PVSUFF\_SUBJ:3MS & * & \\
        PVSUFF\_SUBJ:3FS & * & at \\
        PVSUFF\_SUBJ:1P & * & na \\
        PVSUFF\_SUBJ:\{2D,2MP,2FP\} & * & tu \\
        PVSUFF\_SUBJ:3MP & \<وْا> \textit{woA} & ew \\
        PVSUFF\_SUBJ:\{3MD,3FD,3MP,3FP\} & * & u \\
        CVSUFF\_SUBJ:\{2MS,2FS\} & * & \\
        CVSUFF\_SUBJ:2MP & * & u \\
        PRON\_1S & \<نِي> \textit{niy} & ni \\
        \{PRON,POSS\_PRON\}\_1S & * & i \\
        \{PRON,POSS\_PRON\}\_\{2MS,2FS\} & * & ek \\
        PRON\_3MS & * & u \\
        POSS\_PRON\_3MS & * & u \\
        \{PRON,POSS\_PRON\}\_3FS & * & ha \\
        \{PRON,POSS\_PRON\}\_1P & * & na \\
        \{PRON,POSS\_PRON\}\_\{2D,2MP,2FP\} & * & kom \\
        \{PRON,POSS\_PRON\}\_\{3D,3MP,3FP\} & * & hom \\
        NSUFF\_FEM\_SG & * & a \\
        NSUFF\_MASC\_DU\_* & * & ejn \\
        NSUFF\_FEM\_DU\_* & * & tejn \\
        NSUFF\_MASC\_PL\_* & * & in \\
        NSUFF\_FEM\_PL & * & iet \\
        \bottomrule
    \end{tabular}
    \caption{
        Morpheme Mappings indicating how different morpheme classes are mapped; some tags are collapsed to a single row for presentation purposes.
        When the Arabic form is specified in the Source column, the target mapping is only applied to this specific form, otherwise it applies to all forms (indicated by *).
        `\_' indicates an explicit spacing added so that the morpheme is no longer attached to the word.
    }
    \label{table:morpheme_mappings}
\end{table*}

\section{Data Normalisation}
\label{appendix:data_normalisation}

\paragraph{Named-Entity Recognition (NER)}
Since the MAPA \cite{gianola-etal-2020-mapa} and ANERCorp \cite{benajiba-etal-2007-anersys} datasets have a different tagset, we normalise these to only keep Person (\texttt{PER}), Organisation (\texttt{ORG}), and Location (\texttt{LOC}) tags, dropping every other tag (\texttt{O}).
For ANERCorp, this means that we only removed the Miscellaneous tags (\texttt{MISC}).
However, the MAPA data has more fine-grained annotations.
Hence, for MAPA, we only keep the \texttt{PER} the token's level 2 tag is either a \texttt{given name} or a \texttt{family name}, and we designate \texttt{LOC} for tokens marked as \texttt{city}
or \texttt{country}.
Originally, the MAPA data had 3,901 sentences which is comparable to the ANERCorp data.
Since we are interested in data augmentation under a resource-constrained setting, we choose to downsample the MAPA training data to allow us to better measure this.
Hence, we downsample the Maltese data, so that we have around the same ratio of sentences as the Arabic and Maltese datasets for Sentiment Analysis.
By doing so, we end up with 155 and 43 sentences for training and validation, respectively, and the test set remains unchanged.

\paragraph{Sentiment Analysis (SA)}
The Maltese dataset from \citet{martinez-garcia-etal-2021-evaluating} only has positive or negative labels, whereas the Arabic dataset from \citet{baly-etal-2018-arsentdlev} has positive, negative, or neutral labels.
Hence, we drop all Arabic sentences with a neutral label, ending up with 15,305 sentences.
The Maltese data remains unchanged.

\section{Fine-Tuning Details}
\label{appendix:fine_tuning}

For our experimental setup described in Section~\ref{section:setup}, we fine-tune BERT-based models by adding a linear token or sentence classification head, depending on the task.
We use the Transformers library to conduct all of our experiments \cite{wolf-etal-2020-transformers} and the code is made publicly available.\footnote{
\url{https://github.com/MLRS/BERTu/tree/main/finetune}
}

For all tasks, we use an inverse square root learning rate schedule with a maximum learning rate of 2e-5 and a warmup of 1 epoch.
We also set the classifier dropout to 0.1 and the weight decay to 0.01.
We train with batch sizes of 16 for a maximum of 200 epochs early stopping on the development set\footnote{
The development set used is always the Maltese data, even when fine-tuning with Arabic data.
} with a patience of 20 epochs.
Each experiment is performed 5 times with different random seeds, reporting the average across these runs.

We fine-tune all models on a compute cluster using A100 GPUs.
Fine-tuning runtimes vary largely because we train using early stopping but also depending on the dataset and model used.
On average, a single training run on Arabic data takes around 28 minutes and 44 minutes for Named-Entity Recognition and Sentiment Analysis, respectively, while a single training run on Maltese data takes around 7 minutes and 3 minutes for Named-Entity Recognition and Sentiment Analysis, respectively.

\end{document}